\documentclass{article}

\PassOptionsToPackage{numbers, compress}{natbib}

 \usepackage[preprint]{neurips_2026}

\usepackage[utf8]{inputenc} 
\usepackage[T1]{fontenc}    
\usepackage{hyperref}       
\usepackage{url}            
\usepackage{booktabs}       
\usepackage{amsfonts}       
\usepackage{nicefrac}       
\usepackage{microtype}      
\usepackage{xcolor}         
\usepackage{graphicx} 
\usepackage{amsmath} 
\usepackage{float}
\usepackage{multirow}
\usepackage{caption}
\usepackage[justification=justified,singlelinecheck=false]{caption}
\title{CellDETR:  A Detection-Guided Framework for Scalable Cell Representation Learning from Histopathology Images}

\author{%
  \textbf{Shikang Zhang}\thanks{These authors contributed equally (co-first authors).} \\
  Zhejiang University of Technology \\
  \texttt{\href{mailto:221123030330@zjut.edu.cn}{221123030330@zjut.edu.cn}}
  \And
  \textbf{Guojun Li}\footnotemark[1] \\
  Tianjin University \\
  \texttt{\href{mailto:liguojun.medai@gmail.com}{liguojun.medai@gmail.com}}
  \AND
  \textbf{Yicong Mao} \\
  Zhejiang University of Technology \\
  \texttt{\href{mailto:221124120197@zjut.edu.cn}{221124120197@zjut.edu.cn}}
  \And
  \textbf{Chulin Sha}\thanks{Corresponding author: shachulin@gmail.com} \\
  Hangzhou Institute of Medicine, Chinese Academy of Sciences \\
  \texttt{\href{mailto:shachulin@gmail.com}{shachulin@gmail.com}}
}

\begin{document}

\maketitle
\begin{abstract}
    
    Recent advances in pathology foundation models have substantially improved patch and slide level representation learning from whole-slide images (WSIs). However, cell-level representations learning remain underexplored, limiting cell-resolved interpretability, biological discovery, and clinical translation. We propose CellDETR, a detection-guided framework built on Deformable DETR for scalable cell representation learning from WSIs. By introducing location feature decoupling and box-constrained attention mechanism, CellDETR enables automated extraction of cell-level embeddings, and outperform existing state-of-the-art methods in supervised cell classification on PanNuke data. In addition, by incorporating contrastive learning design, we build a CellDETR-based pretraining model for scalable cell representation learning from unlabeled WSIs, which improves downstream cell classification performance. Furthermore, we show that after pretraining with Xenium spatial transcriptomics-derived cell annotations, CellDETR achieves accurate cross-dataset cell classification, demonstrating the transferability and biological relevance of the learned cell embeddings. Together, CellDETR provides a scalable route toward general cell-level representation learning framework for interpretable computational pathology. 
\end{abstract}

\section{Introduction}
In recent years, the successful application of vision transformer (ViT) \cite{key5} architectures and self-supervised learning (SSL) strategies has led to a paradigm shift in computational pathology. A growing number of pathology foundation models, such as UNI\cite{chen2024uni}, Virchow\cite{vorontsov2024virchow}, and Prov-GigaPath \cite{xu2024gigapath} have been developed using large-scale unlabeled whole-slide images (WSIs). These models can learn robust and transferable representations at patch or slide level, and have shown strong potential in tasks including tumor detection, region-of-interest identification, molecular feature prediction, and prognosis assessment\cite{neidlinger2025benchmarking}\cite{campanella2025clinical}. However, most existing pathology foundation models use fixed-size patches as the basic representation unit. As a result, they do not directly provide reliable and transferable representations at cell level. This limits cell-resolved interpretability from WSIs and restricts their application in tasks such as cell–cell spatial interaction modeling and tumor microenvironment analysis, which are important for both biological discovery and clinical translation.

Cell-level representation learning from WSIs faces two major challenges. The first challenge lies in data availability. High quality cell-level annotations remain limited, and existing manually annotated datasets often contain relatively coarse cell types and restricted tissue diversity\cite{ding2023large}. This makes it difficult to capture the broad spectrum of cellular morphology across tissues, organs, and disease states. The second challenge associates with model design. Currently, most label-free self-supervised models are built upon fixed-size patch inputs and fixed-length token sequences, which are poorly suited to cells with variable size, shape, and number. Existing cell-level analysis methods, such as CellViT \cite{key2} and CellViT++ \cite{key3} have shown that Transformer\cite{vaswani2017attention} encoders can support cell-level feature learning by adding nuclei detection, segmentation, and classification tasks. Nevertheless, these methods are mainly optimized for segmentation and predefined cell-type classification, rather than general cell representation learning. Therefore, how to effectively learn reliable cell-level representations from WSIs that are transferable across datasets and scalable through large pretrain remain largely unexplored.

\begin{figure}
  \centering
  \includegraphics[width=1\linewidth]{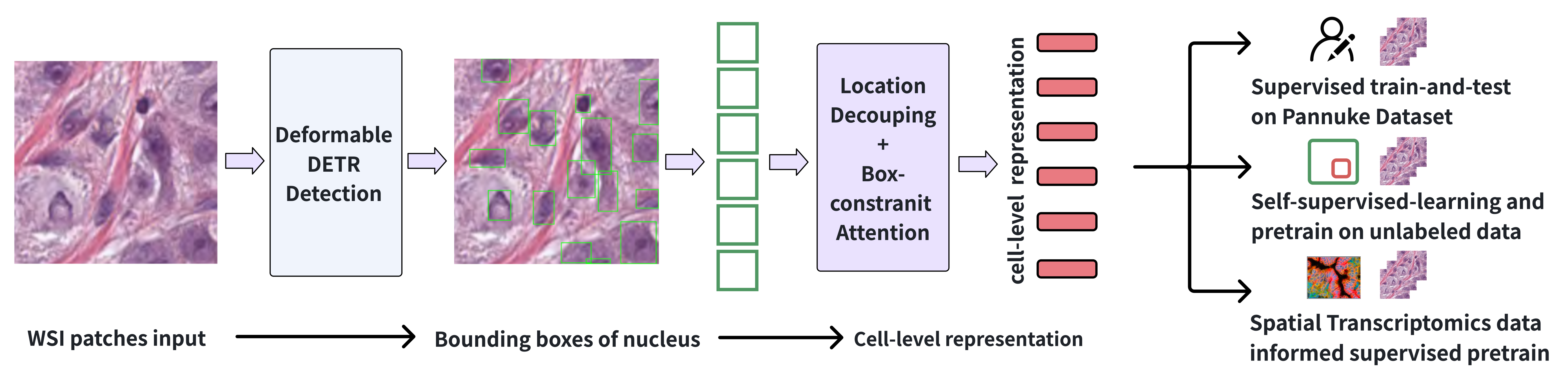}
  \caption{\textbf{Overall Design of Cell DETR framework.} We propose CellDETR, a deformable DETR-based detection-guided framework that extracts transferable cell-level embeddings from pathology images through feature decouple and box-constrained attention design, enabling supervised, self-supervised, and spatial-transcriptomics-informed cell representation learning from WSIs.
}
  \label{fig:idea}
\end{figure}

To address these challenges, we propose CellDETR, a detection-guided framework for cell representation learning based on Deformable DETR\cite{Zhu2021DeformableDETR}. Unlike conventional fixed-patch representation learning, CellDETR first detects a variable number of nuclei within each patch and uses the detected objects as the basic units for cell representation learning (Fig. \ref{fig:idea}). Specifically, we introduce two key modules to strengthen cell-level representation learning (Fig. \ref{fig:model}a-c). First, we decouple localization information from morphology-related features, reducing the influence of positional cues with cell identity representation. Second, we design a box-constrained attention mechanism that guides the model to focus on the corresponding nuclei region when extracting each cell embedding. This reduces contamination from background noise and irrelevant tissue texture. Through this detection-guided design, CellDETR can accommodates the high variability of cells in WSIs, enabling an automated cell extraction and reliable cell-level representation learning.

We first evaluate whether CellDETR can effectively learn cell representation on the PanNuke \cite{gamper2020pannuke} dataset under supervised training. The results show that CellDETR can accurately localize nuclei and generate discriminative cell representations for cell-type classification. Then, we examine whether CellDETR can be extended to SSL setting. We implement this by introducing a contrastive learning strategy and pretrain the model on publicly collected unlabeled WSIs. After fine-tuning on PanNuke, the pretrained model outperforms direct supervised training, demonstrating that CellDETR can learn general cell representations from unlabeled data and is suitable for scalable pretraining. In addition, we used publicly available WSI-paired Xenium\cite{salas2025optimizing} spatial transcriptomics\cite{stahl2016visualization} (ST) data to obtain cell type annotations for CellDETR pretraining. The ST-derived-label pretrained model transfers effectively to PanNuke and achieves good cross-dataset cell classification performance, suggesting that ST data annotations can provide biologically grounded supervision to improve cell representation learning within CellDETR framework.

Together, our work show that CellDETR can learn reliable cell-level representations under supervised, self-supervised, and ST-data-informed training settings  (Fig. \ref{fig:idea}). By combining detection-guided cell-level modeling with flexible pretraining strategies, CellDETR provides a practical route to reduce the dependence on manual cell annotations and scale cell representation learning with unlabeled WSIs, meanwhile supporting the future development of ST-informed cell-level models for interpretable computational pathology. Our code is available at https://github.com/kszstudent/CellDETR.
\vspace{-1em}
\section{Related work}
\subsection{Cell-level Analysis and Cell Representation Learning }

Cell-level analysis is essential for computational pathology. Classical deep learning approaches such as Mask-RCNN\cite{he2017mask}, Micro-Net\cite{raza2019micronet}, and Hover-Net\cite{graham2019hover} have been widely used for nuclei segmentation and classification. Recently, Transformer-based models such as CellViT\cite{key2} and CellViT++ \cite{key3} further improve cell-level analysis.  Despite these advances, most existing methods are optimized for segmentation or predefined cell-type classification, while general cell representation learning remains rather underexplored.

\subsection{DETR and Object Detection Models}

DETR \cite{carion2020end} introduced the Transformer architecture into the field of object detection, achieving end-to-end detection through global self-attention mechanisms. Unlike traditional detection methods that rely on anchors, proposals, or non-maximum suppression, DETR directly predicts a set of object categories and bounding boxes, providing a natural way to model images containing a variable number of objects. Deformable DETR\cite{Zhu2021DeformableDETR} further improves DETR by introducing deformable attention. By combining object queries with adaptive sparse attention, this design accelerates convergence and improves detection of multi-scale and small objects, providing an efficient framework for detecting densely distributed and scale-varying objects. Currently, DETR-based methods have mainly been developed for object localization, and their potential for learning cell-level representations in WSI remains less explored.

\subsection{Self-supervised Learning in Computational Pathology }

To reduce reliance on large-scale manual annotations, self-supervised learning (SSL) has become a key strategy for representation learning in computational pathology. Early SSL methods often rely on contrastive objectives that learn representations by aligning positive pairs and separating negative pairs. Recently, DINO\cite{caron2021emerging} proposed a self-distillation strategy without negative pairs, which has become an important basis for recent pathology foundation models. For instance, UNI\cite{chen2024uni} and Virchow\cite{vorontsov2024virchow} have successfully applied DINO-like self-distillation to large-scale unlabeled datasets, achieving strong generalization across various clinical tasks. Despite their success, existing SSL pathology models mainly learn representations from fixed patch, which are not well aligned with cell representation learning. In this work, we extend DINO-style self-distillation from patch-level to cell-level cell representation learning by matching local and global views of detected object.

\section{Method}
CellDETR is designed to learn cell-level representations from WSIs by treating nuclei rather than fixed patches as basic units. The framework is built upon Deformable DETR\cite{Zhu2021DeformableDETR} which provides an end-to-end object detection architecture for locating variable numbers of nuclei within each patch. However, standard Deformable DETR is not explicitly designed for robust cell representation learning. As the morphological features such as shape, size, and texture are more informative for characterizing cellular identity\cite{heckenbach2022nuclear}, we redesign the decoder of Deformable DETR as a decoupled architecture consisting of a detection decoder and a representation decoder Fig. \ref{fig:model}a. The detection decoder is responsible for nuclei localization, while the representation decoder extracts cell morphological features, thus to separate location information from feature embeddings. More specifically, CellDETR introduces two key modules to strengthen cell-level representation learning: 1) location feature decouple module, and 2) box-constrained attention module. In addition, we also implement SSL pretraining model on unlabeled data by introducing cell-level contrastive learning strategy.

\subsection{Location Feature Decouple Module }
\begin{figure}[H]
  \centering
  \includegraphics[width=1\linewidth]{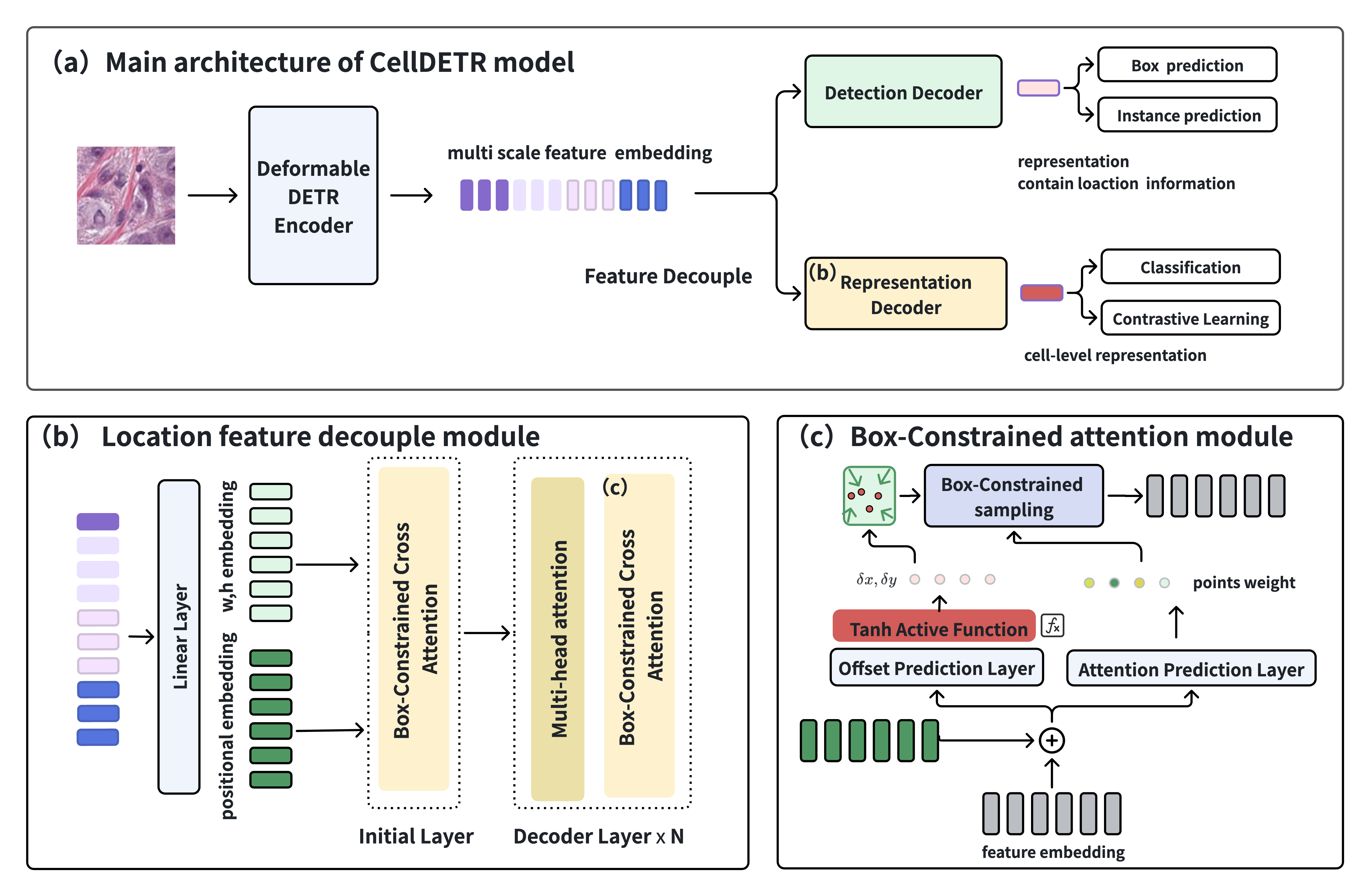}
  \caption{\textbf{The architecture of CellDETR.} (a) The overall model architecture, where the backbone, encoder, and Detection Decoder all follow the standard Deformable DETR structure. (b) Detailed structure of location feature decouple representation decoder, which uses a linear projection and an initialization layer to separate positional information from feature embeddings. (c) Illustration of the box-constrained attention module, where bounding box information and a tanh activation are used to constrain attention within detected cell regions. }
  \label{fig:model}
\end{figure}

The detailed feature decouple is illustrated in this section. We decouple the original Deformable DETR decoder into a detection decoder and a representation decoder. The detection decoder follows the standard Deformable DETR design and predicts nuclei bounding boxes. In the representation decoder, we first introduce an initialization layer inspired by Liu et al.~\cite{liu2022dab}, where the center coordinates $(x,y)$ and box size $(w,h)$ from encoder are separately projected through two linear layers, allowing location and scale information to be encoded in a decoupled manner ( Fig. \ref{fig:model}b).  

Next, we remove the multi-head self-attention layer before deformable cross-attention in the representation decoder. This prevents excessive interaction among different cell queries and encourages each query to extract features mainly from its corresponding cellular region, so as to produce a clean and cell-specific representations.

\subsection{Box-Constrained Attention Module}
The mechanism of deformable attention involves taking a reference point $(x, y)$, adding an offset ($\Delta{x}, \Delta{y}$) predicted by a linear layer, and then performing bilinear interpolation sampling in the multi-scale feature space. The sampling process is formulated as:
\begin{equation}
F_l = x^l\left(\phi_l(\hat{p}_q) + \Delta p_{mlqk}\right)
\label{eq:feature_sampling}
\end{equation}
where the indices $m,l,q,k$ denote the attention head, input feature level, query index and sampling point index, respectively. $\Delta p_{mlqk}$ represents the reference point coordinates normalized to the range [0, 1]. The function  $\phi_l(.)$ is used for scale transformation across different feature levels, and $\Delta p_{mlqk}$ denotes the sampling offset.

Then, the attention linear layer is used to predict the attention weights for the sampled features, followed by a weighted summation to obtain the new representation. This process can be described by the formula:
\begin{equation}
\text{MSDeformAttn}(z_q, \hat{p}_q, \{x^l\}_{l=1}^L)
=
\sum_{m=1}^M
W_m
\left[
\sum_{l=1}^L
\sum_{k=1}^K
A_{mlqk}
\cdot
W'_m F_l
\right]
\label{eq:msdeformattn}
\end{equation}
where $W'_m $ and $W_m$ are the sampling matrix and output projection matrix, $A_{mlqk}$ represents the predicted attention weight, satisfying $\sum_{l=1}^{L} \sum_{k=1}^{K} A_{mlqk} = 1$.

However, in the original deformable attention, the prediction of offsets is not constrained by any region, and the sampling process can be described as: 
\begin{equation}
(\hat{x}, \hat{y})
=
\frac{(\Delta x, \Delta y)}
{n \cdot (w, h) \cdot 0.5}
+
(x, y)
\label{eq:offset_sampling}
\end{equation}
Inspired by the observation in Zhang's work ~\cite{zhang2023decoupled} that localization and classification rely on different attention regions, we introduce a simple box-constrained attention mechanism. Specifically, after the offset prediction linear layer, we apply a tanh activation function (the red background area in Fig. \ref{fig:model}c) and scale the normalized offsets by the width and height of the corresponding bounding box. This restricts the sampling locations to the target box region and encourages the representation decoder to focus on cell-localized features rather than surrounding tissue or neighboring nuclei. The constrained sampling locations are formulated as:
\begin{equation}
(\hat{x}, \hat{y})
=
(x, y)
+
0.5 \cdot
\tanh\left((\Delta x, \Delta y)\right)
\odot
(w, h)
\label{eq:box_constrained_attention}
\end{equation}
where $\odot$ denotes element-wise multiplication.


\subsection{Cell-Level Contrastive Learning for SSL-based Pretraining on Unlabeled Data}
To reduce dependence on manually annotated cell-type labels, we extend DINO-style SSL from patch-level representation learning to cell-level representation learning. Instead of applying contrastive learning to fixed patches, we define self-supervision at the nucleus instance level using bounding boxes. Specifically, for each nucleus, we generate a perturbed box by randomly shifting and scaling its original index box (Fig. \ref{fig:contrstive} a). The perturbed anchor box is used as a local view, while the original box serves as the corresponding global view. The representation decoder is trained to pull together embeddings from the same nucleus (positive pair) and separate them from embeddings of other nuclei (negative pair) in the same, resulting in a cell-level contrastive learning objective.

To prevent information leakage and collapse, we apply an attention mask in the representation decoder: each anchor can attend to itself and all original instance boxes, but not to other anchors. Positional embeddings are also removed from multi-head attention queries to avoid solving the task relying solely on positional cues. The resulting embeddings are passed through a Feed-Forward Network (FFN) layer, and optimized with the Information Noise-Contrastive Estimation (InfoNCE\cite{oord2018representation}) loss. For efficient pretraining, we adopt a parallel training strategy in which padded query slots are replaced with additional anchor boxes, increasing the number of contrastive pairs per batch. This design enables CellDETR to learn morphology-aware cell representations from unlabeled WSIs data.

\section{Experiments}

\subsection{Datasets}

We evaluate CellDETR under supervised, self-supervised, and spatial-transcriptomics-informed settings using three datasets.

PanNuke. We used PanNuke \cite{gamper2020pannuke} as the main supervised benchmark. PanNuke provides manually annotated nuclear boundaries and class labels across five categories: neoplastic, epithelial, inflammatory, connective, and necrotic. We converted nuclear boundaries into bounding boxes and used this dataset to evaluate nuclei detection and cell-type classification, and to compare CellDETR with existing methods.

Unlabeled publicly collected WSI data. For self-supervised contrastive pretraining, we constructed 64,000 HE pathology patches across three tissue types. Each patch has a size of 256×256 pixels and includes nucleus segmentation information, but no cell-type labels. These data were used to generate cell-level instances for contrastive learning and to evaluate whether CellDETR can learn useful cell representations without manual cell annotations.

Xenium-derived spatial transcriptomics and WSI paired dataset. To explore ST-informed cell representation learning, we constructed a dataset of 21,000 patches from public Xenium data. Since Xenium measures gene expression for only a subset of nuclei in each region, this dataset provides partial but more molecularly informed cell-type annotations. We used these labels to train CellDETR and evaluate whether spatial transcriptomics can provide biologically grounded supervision for WSI cell representation learning.

\subsection{Metrics}
We adopt the evaluation metrics recommended in PanNuke \cite{gamper2020pannuke}, including precision, recall, and F1 score for each cell type. For nuclei detection, we also report overall detection precision, recall, and F1 score, where a predicted nucleus is considered correct if it matches a ground-truth nucleus according to the predefined matching criterion in the PanNuke evaluation protocol. 

\subsection{Supervised Evaluation of Cell-Level Representation Learning}

We first evaluate CellDETR on the PanNuke dataset\cite{gamper2020pannuke} under a supervised setting. The model is trained for 100 epochs to jointly perform nucleus detection and cell representation learning. Since the representation decoder can use either predicted boxes from the detection decoder or ground-truth boxes, we design two training strategies (Supplementary \ref{app:training mode}) and compared CellDETR with the standard Deformable DETR baseline, as well as other popular methods including Mask-RCNN, Micro-Net, HoverNet and CellViT. 

As shown in Tab. \ref{tab:supervise}, CellDETR achieves the best overall detection performance on PanNuke, showing its ability to accurately localize nuclei in WSIs. Under the sequential training setting, CellDETR also obtains the highest F1 scores across all cell types, demonstrating that the proposed representation learning design can effectively extract reliable cell-level features for cell-type classification. Notably, CellDETR shows a clear improvement in the classification of necrotic cells, suggesting that cell-level representation learning maybe beneficial for morphologically challenging cell types. Under the parallel training setting, CellDETR maintains performance comparable to the baseline, indicating that the model can learn effective cell embeddings from cell-level box inputs.

\begin{table}[H]
\caption{\textbf{Performance on PanNuke using supervised training.} ``S'' denotes the sequential training mode, ``P'' denotes the parallel training mode. }
\label{tab:supervise}
\centering
\resizebox{\textwidth}{!}{
\begin{tabular}{c|l|lll|lll|lll|lll|lll|lll}
\toprule
& Methods & \multicolumn{3}{c}{Neoplastic} &\multicolumn{3}{c}{Epithelial} &\multicolumn{3}{c}{Inflammatory} & \multicolumn{3}{c}{Connective}  & \multicolumn{3}{c}{Necrotic} & \multicolumn{3}{c}{Detection} \\
\cmidrule(lr){3-5} \cmidrule(lr){6-8} \cmidrule(lr){9-11} \cmidrule(lr){12-14} \cmidrule(lr){15-17} \cmidrule(lr){18-20}
 & & P & R & F & P & R & F & P & R & F & P & R & F & P & R & F & P & R & F \\
\midrule
\multirow[c]{5}{*}{Others}
& Mask-RCNN\cite{he2017mask} & 0.55 & 0.63 & 0.59 & 0.52 & 0.52 & 0.52 & 0.46 & 0.54 & 0.50 & 0.42 & 0.43 & 0.42 & 0.17 & 0.30 & 0.22 & 0.76 & 0.68 & 0.72 \\
& Micro-Net\cite{raza2019micronet} & 0.59 & 0.66 & 0.62 & 0.63 & 0.54 & 0.58 & 0.59 & 0.46 & 0.52 & 0.40 & 0.45 & 0.47 & 0.23 & 0.17 & 0.19 & 0.78 & 0.82 & 0.80  \\
& HoverNet\cite{graham2019hover} & 0.58 & 0.67 & 0.62 & 0.54 & 0.60 & 0.56 & 0.56 & 0.51 & 0.54 & 0.52 & 0.47 & 0.49 & 0.28 & 0.35 & 0.31 & 0.82 & 0.79 & 0.80  \\
& CellViT\cite{key2} & 0.69 & 0.70 & 0.70 & 0.68 & 0.71 & 0.70 & 0.59 & 0.58 & 0.58 & 0.53 & 0.51 & 0.52 & 0.39 & 0.35 & 0.37 & 0.83 & 0.82 & 0.82 \\
& Deformable DETR \cite{pina2024celldetr}& 0.72 & 0.67 & 0.69 & 0.71 & 0.67 & 0.69 & 0.59 & 0.60 & 0.59 & 0.57 & 0.59 & 0.53 & 0.54 & 0.32 & 0.40 & 0.85 & 0.78 & 0.81 \\
\midrule
\multirow[c]{2}{*}{Ours}
& CellDETR-S &0.71&0.70& \textbf{0.70}& 0.70&0.71 &\textbf{0.70} &0.61&0.61 &\textbf{0.61} &0.57 &0.51 &\textbf{0.54} &0.49 &0.40 &\textbf{0.43} & 0.83 & 0.81 &\textbf{0.82}  \\
& CellDETR-P &0.69& 0.68&0.68 & 0.68&0.67 &0.67 &0.58 &0.61 &0.60 & 0.56& 0.50&0.53 &0.50 &0.36 &0.41 &0.83 &0.79 &0.81 \\
\bottomrule
\end{tabular}
}
\end{table} 

\begin{figure}[bh]
  \centering
  \includegraphics[width=1\linewidth]{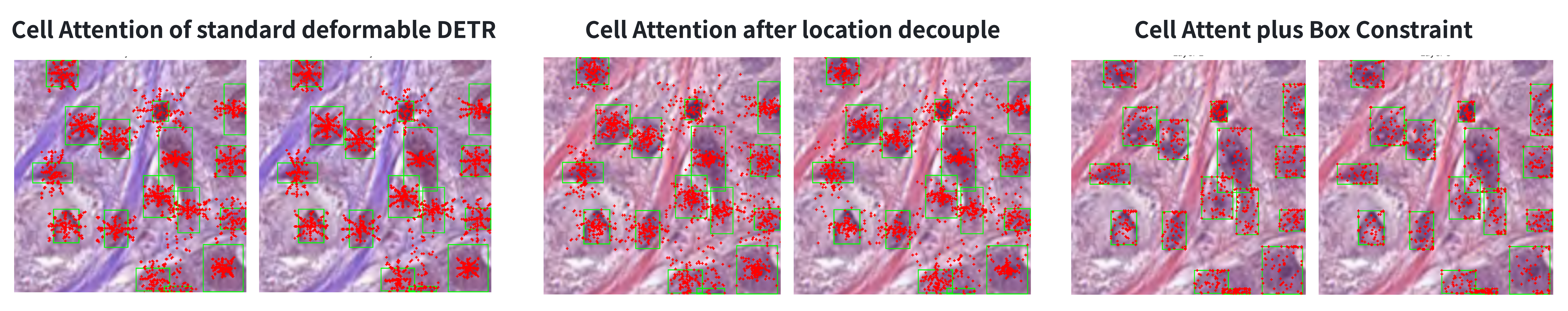}
  \caption{\textbf{Illustration of improved cell-specific feature attention of location decouple module and box-constraint attention module.} We present the attention regions of the last two decoder layers of deformable attention. The attention map shows both location decouple module and box-constraint attention module are effective in cell-specific feature extraction.}
  \label{fig: attention}
\end{figure}

We further examine the contribution of location feature decoupling and box-constrained attention modules. As shown in Fig. \ref{fig: attention}, compared with standard Deformable DETR, CellDETR produces  more accurate attention within cellular regions. The attention maps concentrate more accurately and evenly distributed on target nuclei, with less distraction from surrounding tissue or neighboring cells. These results support the effectiveness of our design in reducing background contamination and improving cell-specific feature extraction, showing that CellDETR learns more reliable and discriminative cell-level representations for downstream analysis.
 
\subsection{Self-Supervised Scalable Pretraining On Unlabeled Data}

To examine whether CellDETR can learn cell-level representations without cell-type annotations, we introduce a DINO-inspired self-supervised pretraining strategy in CellDETR framework (See Methods 3.3 for details). We pretrain the developed model for 30 epochs on 64,000 publicly collected WSIs, using a noise perturbation level of 0.1–0.4 and a temperature of 0.1. And evaluate the performance on PanNuke data under two settings: frozen-feature evaluation and supervised fine-tuning.  

\begin{figure}
  \centering
  \includegraphics[width=1\linewidth]{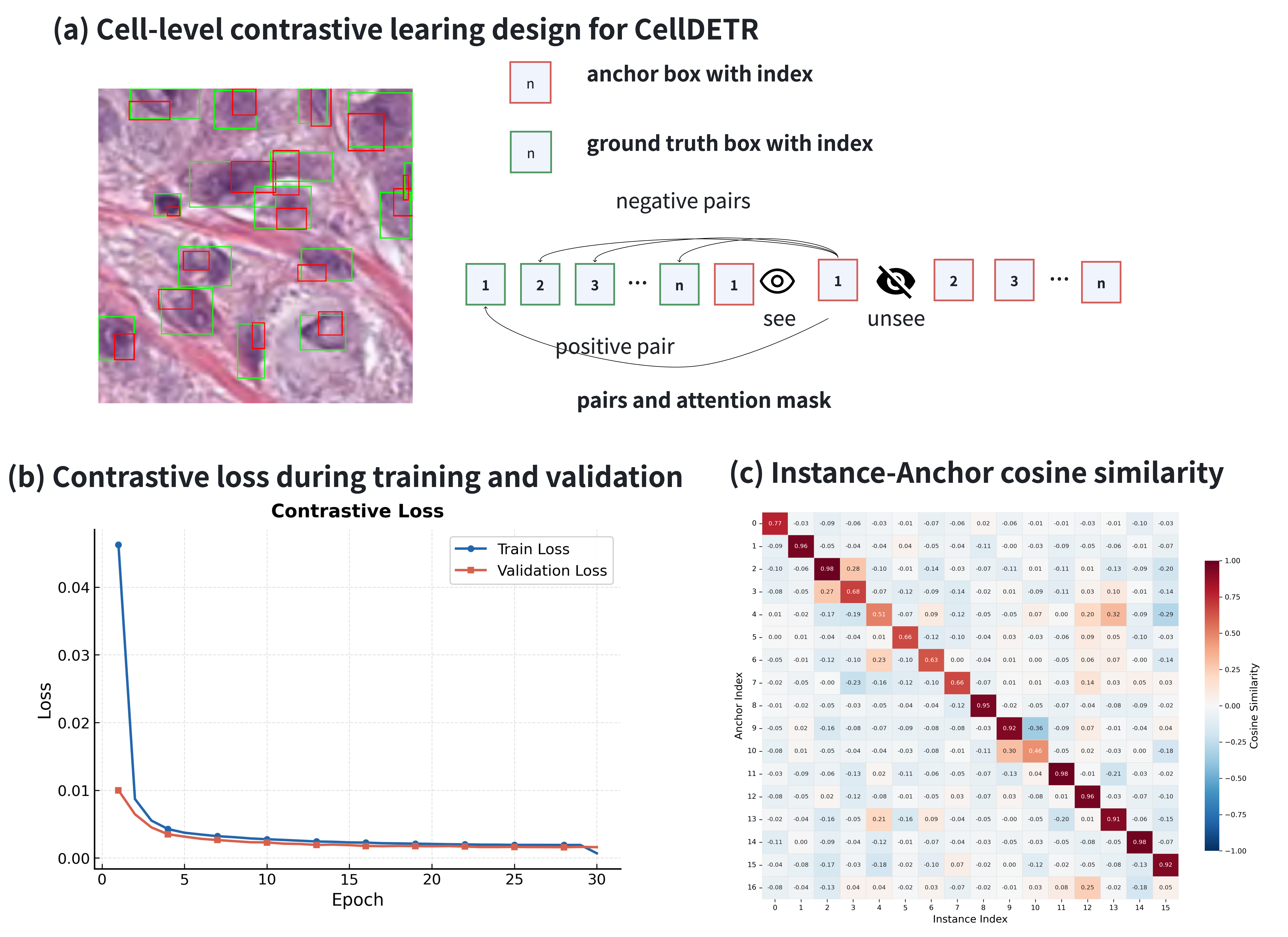}
  \caption{\textbf{Contrastive learning implementation and results.} (a) Illustration of cell-level contrastive learning design for CellDETR. The training section includes the loss descent curve and a similarity heatmap between the anchor and ground truth in the contrastive learning representation space. Both training and validation losses decrease normally, and the high values along the diagonal in the similarity heatmap indicate that the model successfully accomplishes the proxy task. }
  \label{fig:contrstive}
\end{figure}

We first evaluate whether CellDETR could effectively optimize this SSL-based pretraining task. As shown in Fig. \ref{fig:contrstive}b, both training and validation losses decreased steadily during contrastive pretraining, indicating stable optimization. We then measure the cosine similarity between the anchor boxes and the instances representations during the contrastive training process. By epoch 20, the similarity heatmap shows strong matching between each PanNuke anchor and its corresponding instance, while similarities to other instances remained low (Fig. \ref{fig:contrstive}c). This suggests that the model successfully learned instance-level alignment between perturbed local boxes and their corresponding original cell instances.

We next evaluated the transferability of the learned representations on PanNuke nucleus classification. In the frozen-feature setting, CellDETR achieved performance comparable to convolution-based methods, indicating that the pretrained embeddings already contain useful cell-level information. After full-parameter fine-tuning, both detection and classification performance improved further, with the best F1 scores achieved for neoplastic, epithelial, inflammatory, connective cells, and overall detection (Tab. \ref{tab:contrastive}). These results show that self-supervised pretraining enables CellDETR to learn transferable cell-level representations from unlabeled pathology images, supporting its scalability for cell representation learning without manual cell-type annotations.

\begin{table}[H]
\caption{\textbf{Performance of Self-Supervised CellDETR on PanNuke.} We evaluate self-supervised CellDETR under two settings: linear probing, where the backbone is frozen and only a linear classifier is trained, and full fine-tuning, where all model parameters are optimized.}
\label{tab:contrastive}
\centering
\resizebox{\textwidth}{!}{
\begin{tabular}{l|lll|lll|lll|lll|lll|lll}
\toprule
\multirow[c]{2.5}{*}{Fine-tuning setting} & \multicolumn{3}{c}{Neoplastic} &\multicolumn{3}{c}{Epithelial} &\multicolumn{3}{c}{Inflammatory} & \multicolumn{3}{c}{Connective}  & \multicolumn{3}{c}{Necrotic} & \multicolumn{3}{c}{Detection} \\
\cmidrule(lr){2-4} \cmidrule(lr){5-7} \cmidrule(lr){8-10} \cmidrule(lr){11-13} \cmidrule(lr){14-16} \cmidrule(lr){17-19}
 & P & R & F & P & R & F & P & R & F & P & R & F & P & R & F & P & R & F \\
\midrule
Linear probing &0.61&0.59& 0.59& 0.58&0.38 &0.46 &0.55 & 0.50 &0.53 &0.53 &0.39 &0.45 & 0.50 & 0.13 &0.21 &0.86 &0.69 &0.76 \\
Full fine-tuning &0.70& 0.72& \textbf{0.71} & 0.72&0.72 & \textbf{0.72} &0.62 &0.60 &\textbf{0.61}& 0.59& 0.51& \textbf{0.55} &0.48 &0.39 &0.43 & 0.84 & 0.81 & \textbf{0.83}\\
\bottomrule
\end{tabular}
}
\end{table} 
\subsection{Spatial-Transcriptomics-Informed Pretraining}
\begin{figure}[!t]
  \centering
  \includegraphics[width=0.8\linewidth]{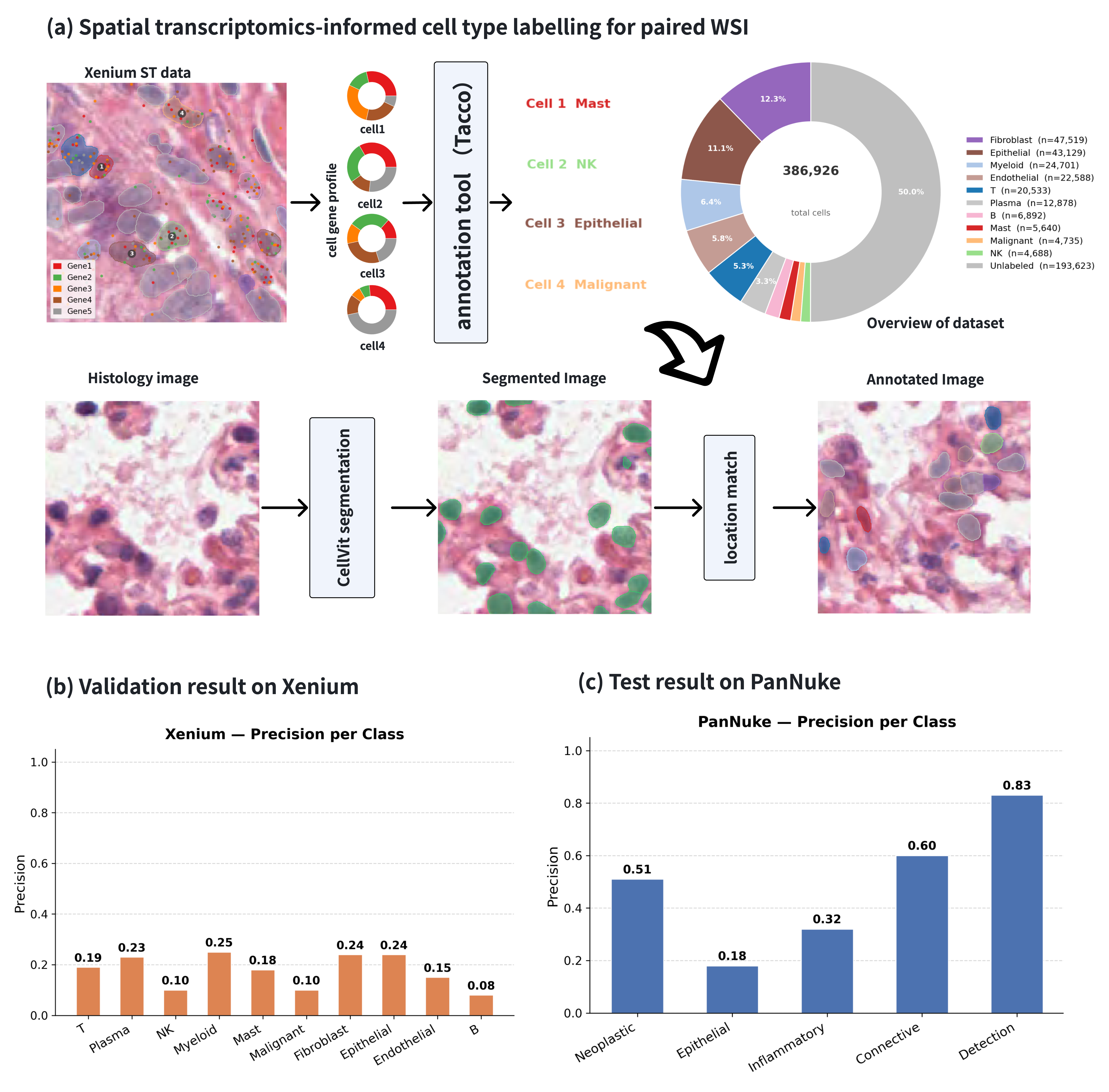}
  \caption{\textbf{Semi-supervised Dataset Construction and Application.}In the data construction phase, we obtain single-cell gene expression profiles from Xenium spatial transcriptomics and derive cell labels using the annotation tool (Tacco\cite{efremova2024tacco}). Nuclear boundaries are segmented from pathological images via CellViT, followed by positional alignment to map cell type information onto the segmentation masks. The annotated cell counts in our dataset are illustrated in a pie chart, showing that approximately half of the nuclei are labeled. The model training and output section displays the results of a pre-trained model that maps cell types into four categories, directly evaluated on PanNuke. The right side presents the results tested on a held-out 20\% portion of the Xenium dataset.}
  \label{fig:semi-supervised}
\end{figure}

We further explored spatial transcriptomics (ST) data as a biologically grounded supervision source for cell representation learning based on CellDETR framework. Using publicly collected WSI-ST paired Xenium data(see details in Datasets), we derived annotations for ten ST-informed cell types on 21,000 patches containing over 380 thousand cells (Fig \ref{fig:semi-supervised}a), and used them to train CellDETR. The model was trained for 20 epochs using the same architecture as in the supervised setting. We first evaluate the Xenium-trained model on the held-out Xenium test set. The model achieved reasonable classification performance on ST-defined cell types, indicating that CellDETR can learn morphology-associated features from transcriptome-derived annotations (Fig \ref{fig:semi-supervised}b). 

To evaluate the transferability to PanNuke, the ten ST-derived cell types were mapped to the major PanNuke categories, and necrotic cells were excluded because they cannot be identified by the ST-derived annotation process. The result shows Zero-shot direct transfer to PanNuke remained challenging: the model showed partial recognition of epithelial and inflammatory cells, while overall performance was limited. This likely reflects label noise, dataset shifts, and imperfect alignment between molecularly defined Xenium labels and morphology-defined PanNuke categories. After further adaptation to PanNuke, Xenium-based pretraining achieved performance comparable to existing supervised state-of-the-art methods (shown in Tab. \ref{tab:weak_and_semi}),  indicating that ST -derived annotations can provide useful biological supervision for transferable cell representation learning. These results support the potential of using ST data to complement morphology-based annotation and may provide a scalable route to reduce reliance on manual cell-level labels.
\begin{table}[H]
\caption{\textbf{Performance of CellDETR on PanNuke with ST-Informed Pretraining.} We evaluate the performance of CellDETR under two settings. In the zero-shot setting, CellDETR trained with ST-informed semi-supervision is directly evaluated on PanNuke without further adaptation. In the fine-tuning setting, CellDETR is further fine-tuned on the PanNuke dataset.
}
\label{tab:weak_and_semi}
\centering
\resizebox{\textwidth}{!}{
\begin{tabular}{l|lll|lll|lll|lll|lll|lll}
\toprule
\multirow{2}{*}{Evaluation setting} & \multicolumn{3}{c}{Neoplastic} &\multicolumn{3}{c}{Epithelial} &\multicolumn{3}{c}{Inflammatory} & \multicolumn{3}{c}{Connective}  & \multicolumn{3}{c}{Necrotic} & \multicolumn{3}{c}{Detection} \\
\cmidrule(lr){2-4} \cmidrule(lr){5-7} \cmidrule(lr){8-10} \cmidrule(lr){11-13} \cmidrule(lr){14-16} \cmidrule(lr){17-19}
 & P & R & F & P & R & F & P & R & F & P & R & F & P & R & F & P & R & F \\
\midrule
Zero Shot &0.51&0.11& 0.02& 0.18&0.43 &0.26 &0.32 & 0.39 &0.35 &0.60 &0.16 &0.26 & - & - & - &0.83 &0.41 &0.55 \\
Finetune &0.69& 0.68& \textbf{0.69} & 0.71&0.62 & \textbf{0.67} &0.59 &0.61 &\textbf{0.60}& 0.60& 0.46& 0.52 &0.48 &0.36 &\textbf{0.42} & 0.84 & 0.79 & 0.81\\
\bottomrule
\end{tabular}
}
\end{table}

\section{Conclusion}

In summarize, we proposed CellDETR, a detection-guided framework for cell-level representation learning from WSIs. By modeling nuclei as basic units, and introducing location-feature decoupling with box-constrained attention mechanism, CellDETR addresses the mismatch between fixed patch representations and variable cellular features, enabling effectively extract reliable cell embeddings from WSI patches. Across supervised train-and-test, self-supervised pretrain and spatial transcriptomics data informed pretrain experimental settings, CellDETR demonstrates a flexible and effective framework for cell-level representation learning. Under supervised training, it improves nuclei detection and cell-type classification on PanNuke, validating the benefit of detection-guided object-level modeling. With DINO-inspired self-supervised pretraining, CellDETR learns useful cell representations from unlabeled pathology images, showing its scalability beyond manually annotated datasets. With ST-derived annotations, the framework further incorporates molecularly informed supervision, suggesting its potential to connect H\&E morphology with biologically defined cell identities. Together, these results show that CellDETR can support supervised, self-supervised, and spatial-omics-informed training within a unified framework, making it a promising step toward scalable and transferable cell-level pathology models.

Nonetheless, this study has several limitations. The current evaluation is based on relatively limited datasets, and although the improvements are consistent, the performance gains remain moderate in some settings. The alignment between ST-derived labels and morphology-based cell categories is also imperfect, which may affect cross-dataset evaluation. Future work will involve larger multi-center cohorts, broader tissue and cancer types, and more paired WSI–ST data. Further integration of cell embeddings with spatial neighborhood modeling may also enable more interpretable analysis of tumor microenvironment organization and cell–cell interactions. Overall, CellDETR provides a practical step toward scalable, transferable, and spatial-omics-informed cell-level foundation models for computational pathology.
\medskip
\bibliography{cite}
\bibliographystyle{unsrtnat}
\appendix

\section{Supplementary}
\subsection{Two training mode}
\label{app:training mode}
In the \textbf{sequential training} strategy, predicted boxes from the detection decoder are directly passed to the representation decoder, allowing the model to learn detection and representation in an end-to-end manner. In the \textbf{parallel training} strategy, annotated boxes are used as inputs to the representation decoder during training, while predicted boxes are used during inference, enabling us to examine the influence of box quality on representation learning.

\begin{itemize}

\item \textbf{ Sequential Mode:} During training, all bounding boxes predicted by the detection encoder are directly passed to the task decoder. The task decoder then performs category prediction only for the instances identified by the detection decoder. During inference, the final detection decision is determined by integrating the confidence scores from both the task decoder and the detection decoder.

\item \textbf{Parallel Mode:} During training, the input bounding boxes for the task decoder are sourced from ground-truth annotations in the dataset. To address the issue of inconsistent numbers of instances per image within a batch, we adopt the maximum number of instances across the batch and pad the remaining entries with dummy bounding boxes set to (0.5, 0.5, 1, 1). A subset of these padding boxes can optionally be designated as $cls_{box}$ tokens to incorporate global contextual information. During inference, only the bounding boxes predicted as instances by the detection decoder are passed to the task encoder, following the same padding strategy. This reduces the sequence length processed by the task encoder to the order of the number of actual instances. In contrast, standard DETR architectures, which often require a fixed sequence length exceeding 400 to ensure detection accuracy, benefit significantly from this design: it accelerates both training and inference speeds.

\end{itemize}
\subsection{Noise adding}
\label{app:noise}
We can describe the formula for generating contrastive learning noise as follows:\begin{equation}
    \begin{aligned}
    \text{mask}_w &\sim \text{Bernoulli}(\rho) \\
    \text{scale}_{w,h} &= 
    \begin{cases}
        u_1 \sim U[s_l, s_u], & \text{if } \text{mask}_w = 1 \\
        u_2 \sim U[-s_u, -s_l], & \text{if } \text{mask}_w = 0
    \end{cases} \\
    \tilde{w} &= w \cdot (1 + \text{scale}_{w,h}) \\
    \tilde{h} &= h \cdot (1 + \text{scale}_{w,h})
\end{aligned}
\end{equation}
\begin{equation}
\begin{aligned}
    \text{mask}_c &\sim \text{Bernoulli}(0.5) \\
    \text{scale}_{c_x,c_y} &= 
    \begin{cases}
        u_3 \sim U[s_l, s_u], & \text{if } \text{mask}_c = 1 \\
        u_4 \sim U[-s_u, -s_l], & \text{if } \text{mask}_c = 0
    \end{cases} \\
    \tilde{c}_x &= c_x + 0.5 w \cdot \text{scale}_{c_x} \\
    \tilde{c}_y &= c_y + 0.5 h \cdot \text{scale}_{c_y}
\end{aligned}
\end{equation}
Equation (1) defines the size scaling noise: first, a binary mask$\text{mask}_w$ is sampled from a Bernoulli distribution to control the scaling direction (the parameter $\rho$ determines the probability of taking 1 for enlargement or 0 for reduction). The scaling ratio is obtained by uniformly sampling from a preset scaling range  [ $s_l$, $s_u$]. The final noisy box width $w'$ and height $h'$ are calculated by multiplying the original box dimensions by (1+scaling$*$ratio). Equation (2) defines the center position perturbation noise, following a process similar to size scaling. The binary mask $\text{mask}_c$ controlling the perturbation direction has its 0/1 probabilities fixed at 0.5, ensuring equal likelihood of perturbation to the left/right or up/down. The perturbation ratio is sampled uniformly in a preset range  [ $s_l$, $s_u$] It is important to note that the actual offset of the center point is determined by the dimensions of the box, i.e., offset = (box$x$width/2)×perturbation$x$ ratio and (box $x$ height/2)×perturbation$x$ratio. This design ensures that the perturbed box center remains approximately within the original box, thereby preventing a complete loss of spatial correlation between the noisy box and the real instance.






\newpage

 \end{document}